\documentclass[a4paper]{article}

\usepackage[english]{babel}
\usepackage[utf8x]{inputenc}
\usepackage[T1]{fontenc}

\usepackage[a4paper,top=3cm,bottom=2cm,left=3cm,right=3cm,marginparwidth=1.75cm]{geometry}

\usepackage{amsmath}
\usepackage{graphicx}
\usepackage[colorinlistoftodos]{todonotes}
\usepackage[colorlinks=true, allcolors=blue]{hyperref}

\usepackage{graphicx}
\usepackage{hyperref}
\usepackage{multicol}

\usepackage{subfigure}

\usepackage{amsmath,array,graphicx}
\usepackage{kantlipsum}
\usepackage{booktabs,lipsum}

\usepackage{setspace}

\usepackage{graphicx}
\usepackage[english]{babel}
\usepackage[utf8x]{inputenc}

\usepackage{lmodern}

\usepackage{amsmath}

\usepackage{amsmath}
\usepackage{amssymb}
\usepackage{fixltx2e}
\usepackage{textcomp}
\usepackage{mathptmx}

\usepackage{booktabs}

\usepackage{flexisym}

\usepackage{mathtools}

\usepackage{float}

\usepackage{algorithm}
\usepackage[noend]{algpseudocode}

\usepackage{type1cm}  
\usepackage{stackengine}
\usepackage{mathptmx}     

\newcommand{\be}{\begin{definition}}
\newcommand{\ee}{\end{definition}}

\title{A Bayesian and Machine Learning approach to estimating Influence Model parameters for IM-RO}
\author{Trisha Lawrence\\ 
Department of Mathematics and 
Statistics\\University of 
Saskatchewan \\
106 Wiggins Road\\ Saskatoon, SK S7N 5E6, CANADA}

\begin{document}
\maketitle

\begin{abstract}

The rise of Online Social 
Networks (OSNs) has 
caused an insurmountable amount 
of interest from advertisers and researchers 
seeking to monopolize on its 
features. Researchers aim to 
develop strategies for 
determining how information is 
propagated among users within  
an 
OSN that is captured by diffusion or 
influence models.
We consider the influence models 
for the IM-RO problem, 
a  novel formulation to the  
Influence Maximization (IM) 
problem based on 
implementing 
Stochastic Dynamic Programming 
(SDP). In 
contrast to existing approaches 
involving influence spread and 
the theory of submodular 
functions, 
the SDP method focuses on 
optimizing clicks and ultimately 
revenue to advertisers in
OSNs. Existing approaches to
influence maximization have been 
actively researched over the 
past decade, with applications to 
multiple fields, however, our 
approach is a more practical 
variant to the original IM 
problem. In this paper, we 
provide an 
analysis on 
the influence models of the IM-
RO problem by conducting 
experiments on synthetic and 
real-world datasets. We propose 
a Bayesian and Machine Learning 
approach for 
estimating the parameters  
of the influence models for the (Influence Maximization- Revenue Optimization) 
IM-RO 
problem. We present a Bayesian 
hierarchical model and
implement the well-known Naive 
Bayes classifier (NBC), 
Decision Trees 
classifier (DTC) and Random 
Forest 
classifier (RFC) on 
three real-world datasets. 
Compared to previous 
approaches to estimating 
influence 
model parameters, our strategy 
has 
the great advantage of being 
directly implementable in standard 
software packages such as 
WinBUGS/OpenBUGS/JAGS and Apache 
Spark. We demonstrate the 
efficiency and usability of our 
methods in terms of spreading 
information and generating 
revenue for advertisers in the context of OSNs.

\end{abstract}

\section{Introduction}

\label{intro}

OSNs 
possess features that enable 
them to be an effective platform 
for spreading information and 
advertising products. Viral 
marketing through OSNs 
has become an effective means by 
which advertising companies 
monopolize their revenue. For 
example, in 
2016, Twitter's advertising 
revenue totaled \$545 million, 
an increase in 60 \% year-over-
year \cite{Twitter}. This 
phenomenon has led researchers 
and inventors 
to improve and develop 
advertising strategies which  generate high revenue. 
The 
IM problem, formally defined in 
\cite{kempe} as choosing a good 
initial set of 
nodes to target
in the context of influence 
models, has been actively 
researched over the past decade 
with its emphasis on social 
networks and marketing products. 
In \cite{Lawrence}, Hosein and 
Lawrence introduced a SDP model for 
the IM problem and recently in \cite{Lawrence2}, 
this approach was formally
defined as the
IM-RO problem.
The SDP approach diverted 
from previous approaches to 
influence maximization that have been based on 
the theory of submodular 
functions and adopted a novel 
and practical 
decision-making perspective. In this SDP approach, an 
online user clicking on 
an impression or advertising 
link was equated to purchasing a 
product and thus the research focused on
maximizing clicks and ultimately 
revenue to the advertiser \cite{Lawrence,Lawrence2}.
In \cite{Lawrence2}, the SDP 
method for the IM-RO problem was 
demonstrated to generate 
lucrative gains to advertisers; 
causing over 
an  80\%  increase in the 
expected number of 
clicks
when evaluated on various 
networks. 
In this paper, our interests lie 
in the influence models for the 
IM-RO problem and how their 
parameters affect revenue 
optimization.\newline\indent
Influence models are defined by 
node 
and edge probabilities that 
capture 
real-world propagations or the 
spread 
of information amongst users 
within 
a network. Although influence 
models for the IM problem have 
been proposed in 
\cite{Dom,Rich,kempe,Galhotra,GoyalL,Cao,Chakrabarti},  
relatively few researchers have 
investigated methods for 
determining 
their parameters 
\cite{Dom,Saito,Cao,GoyalL}.
Compared to the limited work that 
has been done our proposed methods 
have the great advantage to be
easily implementable in the 
standard BUGS
(Bayesian inference Using Gibbs 
Sampling) and Apache Spark software. Consequently, avoiding the burden
of implementing specific algorithms and possible coding errors.
The 
goal of this paper is to provide 
efficient and easily implementable 
methods for 
determining the
parameters of the Graph Influence 
Model (GIM) and Negative Influence 
Model (NIM) mentioned in \cite{Lawrence2}.\newline\indent
From the work in \cite{GoyalL}, 
three types of 
influence models were classified for the IM problem; 
static models, 
continuous models 
and discrete time models. Influence models have also be 
classified as dependent on 
model parameters or on some constants. For example, the 
Weighted Cascade model in \cite{kempe} and Trivalency 
model in \cite{Chen} estimated  $p_{u,v}$, the parameter 
representing the edge 
probability between node $u$ and 
node $v$  by randomly 
selecting a probability from the 
following 
set $\left\{0.1, 0.01, 0.001 
\right\}$ corresponding to
low, high and medium 
probabilities 
of influence. In \cite{Saito}, 
the authors propose an EM algorithm 
to obtain $\kappa_{v,w}$, the 
diffusion probability through link $(v,w)$ in the Independent Cascade model whilst 
the authors in \cite{Cao} proposed a 
weighted sampling algorithm to 
determine $\theta_{u} s$, the set of 
threshold values under the Linear 
Threshold model.\newline\indent 
The significance and novelty of this paper lies in a novel decision-making 
perspective towards influence 
maximization, defined as the IM-RO 
problem in \cite{Lawrence2}. This 
perspective is 
achieved through implementing SDP, a method primarily used 
in shortest paths and resource 
allocation 
problems 
\cite{BertsekasT,Levi,Nascimentop,Powell}. Because of the significant gains achieved from 
implementing the SDP method, we 
propose influence models to 
further leverage on this 
property. We provide an analysis  on the influence 
models for the IM-RO problem 
namely, 
the GIM and NIM and explore how 
their parameters affect the 
optimal expected number 
of clicks generated under the SDP method and Lawrence Degree Heuristic (LDH) 
proposed 
in \cite{Lawrence2}. This analysis
enables us to identify suitable 
priors for the parameter of interest $\alpha$ in our
Bayesian analysis.
\newline\indent 
Our work is a novel and 
practical 
variant of the original IM 
problem proposed by Kempe et al. in 
\cite{kempe}. The 
IM problem uses diffusion or 
influence models and focuses on 
finding a good set of nodes in 
order to create the maximum
cascade or spread over the 
entire network. 
Though an interesting concept, 
our framework captures a more 
realistic representation of how 
users influence each other within 
an online network.\newline\indent
Previous work has provided 
formal 
ways of modeling the
probability of a user buying a 
product based on his/her 
friends buying the product 
\cite{Dom,Rich,kempe,GoyalL}. 
Similarly, we 
employ the GIM and NIM to 
capture these probabilities 
and adopt a Bayesian and 
Machine 
Learning analysis to determine 
their parameters.
Our proposed methods have the  
advantage of being easily 
implementable in the standard 
BUGS (Bayesian inference Using 
Gibbs Sampling) and Apache 
Spark  softwares. We introduce a  
Bayesian hierarchical 
model to provide a point 
estimate for the parameter of 
interest, $\alpha$, of 
the GIM
by the mean  of the 
posterior distribution. In addition, we 
present and compare the NBC, DTC 
and RFC to learn and predict the 
parameter, $p_{0}$, a user's 
initial probability of 
purchasing a product in the 
absence of influence from 
friends.

\section{Related Work}

Because the IM-RO problem was 
recently 
defined,
the only influence models for 
IM-RO 
problem to date are the GIM and NIM \cite{Lawrence2} .
However, studies have 
been conducted on the diffusion 
or influence models for the IM 
problem in \cite{Dom,Rich,GoyalL}. 
 In \cite{Dom}, 
the authors used a non linear 
model that described the network as 
a Markov 
random field where the 
probability of the $i-th$ 
customer purchasing a product  
depended on the neighbours of 
the customer, the product itself 
and a marketing action offered 
to the customer. They showed 
that these probabilities could 
be obtained using a 
continuous relaxation labeling 
algorithm found in  
\cite{Pelkowitz} and Gibbs 
sampling \cite{Geman}. Our Bayesian 
analysis differs from the approach in \cite{Dom} 
because it is easily 
implementable in the standard BUGS (Bayesian Inference Using Gibbs Sampling), consequently, avoiding the burden
of implementing a specific Gibbs algorithm and possible coding errors. The Bayesian
model also
has the great advantage of 
directly providing an estimate
for the uncertainty in the 
parameters such as credible 
intervals. In addition to this, the work in \cite{Dom,Rich} is  restricted to collaborative 
filtering systems while our 
research is suited to users 
within any OSN. \newline\indent
The authors in
\cite{Dom,Rich,GoyalL} proposed a machine learning approach to learn the parameters of their influence models. In \cite{Rich,Dom}, 
the authors assume a 
naive Bayes model \cite{DomingosP} 
and determine a customer's 
internal probability of purchasing 
a product by simply counting.
Similarly, a 
the machine learning approach 
 is adopted in this paper and  in \cite{GoyalL}.  In 
\cite{GoyalL} the authors 
proposed several influence models and developed machine leaning algorithms for learning
the model parameters and making 
predictions. Their algorithms 
generally took no more than two 
scans to learn the parameters of 
their influence model however our 
implementation of machine learning 
algorithms is achieved much faster 
through Apache Spark, a framework 
designed to fulfill the 
computational 
requirements of massive data 
analysis, and manage the
required algorithms \cite{Salman}. Apache Spark has another advantage of offering 
a 
single framework for processing 
data applications such as the
machine learning algorithms 
used in this paper and can be used with applications in both 
static data and streaming data. 
\newline\indent 
The remainder of this paper is 
organized as follows. We
begin by presenting the GIM and 
NIM for the IM-RO 
problem in Section (\ref{influence 
models}). We introduce the methods 
for estimating the parameters of 
the GIM in Section 
(\ref{methods}). Section 
(\ref{experiments}) provides 
experimental results for our 
methods on synthetic and real-world OSNs.
We conclude the paper
in Section (\ref{conclusion}) by 
summarizing the main 
contributions 
and providing directions for future work.

\section{Influence Model for IM-RO}
\label{influence models}

\subsection{Graph Influence Model}
\label{formalGIM}

The Graph Influence model is 
inspired by the IC model in 
\cite{kempe} and as 
its name suggest, is 
greatly affected by the graphical
structure of the network. The model is given 
by: 
\begin{equation}
p_k[i] = max[0, min[1, p_{0k} + 
(1-(1-\alpha \frac{y}{f})^{f}
)]]\;
\end{equation}

where $p_{0k}$ represents a 
user's initial probability 
of clicking on an impression at 
the start of 
stage $k$, with $p_{0k} = p_{0}$ 
when $k=0$.  $\alpha$ is 
an influence constant and $y$ 
represents the number of users 
given impressions and have
clicked on them. The GIM's reliance 
on the network 
structure stems from the parameter 
$f$ which 
represents the 
number of friends of user $i$ and 
is the value for which 
a user's probability is being raised. In 
these experiments, we investigated a 
range of values for  $\alpha$  both less than 1 and 
greater than 1 to determine its effect on the optimal expected 
number of clicks.

\subsection{Negative Influence Model}
\label{subsec:2}

The NIM supports the same 
parameters as 
the GIM with the addition of the 
negative influence parameters 
$n$ and 
$\beta$.

\begin{equation}\label{rule}
p_k[i] = \max[0, \min[1, p_{0k} + \alpha 
\frac{y}{f} - \beta \frac{n}{f}]]
\end{equation}

Here, $\beta$ generally takes on 
values between 0 and 1 and $n$ 
represents the number of users 
given impressions in stage $k$ 
that have not clicked on 
them. In reality it does not 
make sense to provide a user 
with negative information 
(friends who have not clicked on 
impressions) as the goal is to 
encourage users  to make 
purchases. However, our aim is to understand the effect of different influence models for the IM-RO problem.
Influence models incorporating 
the natural 
behavior of users having a 
negative influence on 
their friends have also been 
presented in  ~\cite{ChenandCollins,Bhagat}.

\section{Methods}
\label{methods}

\subsection{Bayesian Analysis}

\subsubsection{The Bayesian Hierarchical Model}

\begin{figure}[H]
\centering
  \includegraphics[width= 5cm,height=4.3cm]{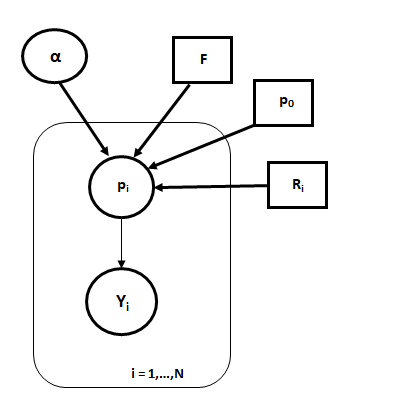}
\caption{Graphical representation with 
plate of Bayesian Hierarchical
Model for $i=1,...,N$ POSTIDs. Rectangular
nodes denote known constants, 
round nodes denote deterministic 
relationships
or stochastic quantities. 
Stochastic dependence is 
represented by
single-edged arrows and 
deterministic dependence is 
denoted by double-edged
arrows}
\label{fig:platediagram}       
\end{figure}

Let $Y_{i}$ $\epsilon$ 
$\mathcal{D}$
represent the responses (number 
of reposts) for a POSTID, $i = 
1,...,N$ and defined by the 
distribution of the data below. The 
probability model for reposting 
a post is represented by the parameters,
$p_{0}$, the initial 
probability of reposting, 
$R_{i}$ the number of times
POSTID $i$ is reposted, $F$, 
the average number of friends
associated with a particular post and 
$\alpha$ the influence constant under the GIM model.
Figure (\ref{fig:platediagram}) 
depicts a graphical 
representation of the 
Bayesian hierarchical model following
\cite{Lunn}.The model 
is as follows:

\begin{equation}\label{bernouilli}
Y_{i}| p_{i} \sim Bernoulli(p_{i})
\end{equation}
with
\[p_{i}= min ( 1, \hspace{0.1cm} 
p_{0}  +  (1 - (1- \alpha * 
\frac{R_{i}}{F})^{F}))\]

A suitable choice of a prior for
$\alpha$ is determined from the 
results of the 
Performance Analysis conducted 
in Section 5 of 
\cite{Lawrence2}. Values of 
$\alpha = 5$ generated the 
optimal expected number of 
clicks on some networks while $\alpha =10$ generated the optimal expected number of clicks on other networks.  Thus, we deduce 
that the network structure also 
influences how 
$\alpha$ affects optimal expected click values.
Therefore, we choose the following 
uniform priors 
for $\alpha$:

\[\alpha \sim Uniform(0,5)\]
and 

\[\alpha \sim Uniform(0,15)\]

\subsubsection{MCMC method}

Monte Carlo Markov Chain methods 
are applied in very complicated 
situations when the data and the 
parameter of interest, say $\theta$ are 
very high dimensional.
Combining the likelihood defined 
by the distribution of the data 
in Equation \ref{bernouilli} and 
the prior gives the joint 
posterior
distribution. Although no 
closed-form expressions exist 
for the posterior distributions, 
simulated values from the 
posterior can
be obtained using a Gibbs 
sampler. 
The method is described as 
follows: 
\newline\indent 
Suppose $\theta$ is our 
parameter of 
interest : $\theta$ = 
($\theta_{1}$,...,$\theta_{d}$) 
$\in$ $\Theta$ $\subseteq$ 
$\mathbb{R}$. We know that 

\[\pi (\theta | X = x) \propto \pi(\theta) f(x;\theta) \]

but there is no practical method 
of computing the normalizing 
constant to make this into a 
proper density function. 
Therefore, we generate a 
pseudo random sample of 
observations from 
$\pi$($\cdot$ $\mid$ x ),
sampling from the distribution 
of $\theta$, holding $x$ 
fixed. Then we can easily 
approximate statistics and 
probabilities of 
interests.
Because a posterior distribution 
is available for all of the 
parameters, a posterior 
distribution is also available 
for $\alpha$.
Hence, JAGS \cite{Plummer} a
software using 
the BUGS syntax is used to 
specify the Bayesian model, by 
drawing random numbers to 
simulate a sample from  the 
posterior to 
form the probability density.
The results for this experiment 
are discussed in Section \ref{experiments}.

\subsection{Machine Learning Algorithms}
\label{sec:3}

For the Machine Learning analysis, we provide 
description of classification 
algorithms implemented to learn the 
mapping from inputs or 
feature vectors, 
$\vec{z}$ to the special feature 
set known as the 
class label, $y$ where $y\in \{1,2,3,...,C\}$ 
and $C$ represents the number of 
classes.

\subsubsection{Naive Bayes}

For the Naive Bayes classifier 
(NBC), the model is  
derived from Baye's theorem which 
states:

\begin{align*}
P(Z=z |Y = y) &= \frac{P(Z=z, Y=y)}
{P(Y=y)}\\
&=\frac{P(Z=z)P(Y=y|Z=z)}
{\sum_{z^'}P(Z=z^')P(Y=y|Z=z)}\\
\end{align*}

where $Y$ and  $Z$ are two random 
variables and the process is 
implemented in two 
steps.\newline\indent
For the first step, the process 
involves 
learning the classification from a 
training dataset $D$ which comprises of features whose class 
labels are known. The  classifier given by:

\begin{equation}\label{NBC}
p(\vec{z}|y = c, \theta) =\prod _{j=1} ^D p(z_j |y = c, \theta _{jc})
\end{equation}

learns the class-conditional 
probabilities 
$P(Z_{i}=z_{i}| y=c)$ of each feature $z_{i}$ 
given the 
class label $c$. Equation $(\ref{NBC})$ 
hinges on the Naive Bayes assumption that the 
features are conditionally independent given the 
class label \cite{Witten}. After learning the 
classifier, the second step, the predicting of  
the posterior probability of the classes is 
given by the NBC prediction model:

\begin{equation}\label{NBC predict}
P(y=c | \vec{z},D) 	\propto p(y=c|D)\prod_{j=1}^D 
p(z_{j}| y=c, D)
\end{equation}

An estimate $\hat{\pi}(c)$, the 
MLE for class $c$ is calculated 
by counting as:

\begin{equation*}
\hat{\pi}(c) = \frac{N_{c}}{N}
\end{equation*}

where  $N_{c}$ is the total number 
of samples in class $c$ and 
$N$ is the total number of samples. 
The implementation for these 
experiments is executed 
through 
Spark Mlib \cite{Meng} with the 
Scala version (2.1.0) which supports a 
multinomial Naive Bayes as its default 
model parameter.

\subsubsection{Decision Trees}

A DTC classifier comprises of a 
hierarchical structure of nodes and directed edges 
which achieves classification by asking a series of 
questions. Although they are easy to implement and are considered 
more informative since they can readily identify significant 
attributes for further analysis \cite{Quilan}, they are prone to 
overfitting. Thus an ensemble of trees tend to generate more accurate 
results \cite{Bauer,Dietterich}. The DTC algorithm  can be 
summarized into the 
following two broad steps:

\begin{enumerate}
 
\item Let $D_{r}=  \left\{(z_{i}, y_{i}),...  \right\}$ be the set of training data  
belonging 
to node 
$r$. At each internal 
node, predictions  $p(y=1|z_{i})$ are 
made over class labels conditioned on 
features and the question is asked `is the 
feature 
$z_{i} \leq t_{i}$', where $t_{i}$ 
is 
a threshold value. The answer to 
this question is a binary variable 
and corresponds to a descendant 
node.

\item After the descendant nodes are created based on 
each outcome,
the samples in $D_{r}$ are then 
distributed to each appropriate descendant node 
based on the response outcome. The 
algorithm continues recursively for 
each descendant node until all of the data is classified.
\newline\indent
The size of the decision tree is 
crucial to the decision tree model 
since too a large a decision tree 
results in over-fitting and too 
small a decision tree results in 
high misclassification rates. Upon 
implementing DTC, it is common to 
grow a tree large enough and prune 
the tree with a set of pruning 
rules found in \cite{Mitchell}. 
However, for 
these experiments, the maximum depth 
of the tree was set to be 5 and  N-
fold cross validation was executed 
to  select and evaluate the best 
decision tree model under a suitable 
metric.

\end{enumerate}

\subsubsection{Random Forests}

RFC was first introduced in 
\cite{Breiman}. The method involves growing ensembles of  
decision 
tree predictors in which
each node is split using the best 
among a subset
of predictors, which are randomly 
chosen at that 
particular node. There are 
numerous advantages to 
implementing 
RFC algorithms as indicated in 
\cite{Horning}. They are robust 
against over-fitting, less 
sensitive to outlier data and have 
high prediction.
The basis steps of a RFC 
classification algorithm are 
summarized as follows:

\begin{enumerate}

\item Given a training set,$ \left\{(z_{i}, y_{i}),...  \right\}$, sample 
$D_{n}$ a set of bootstrap samples where $n$ corresponds to the number of trees.

\item For each of the samples, grow or train a decision 
classification 
tree $f(\vec{z}, D_{n})$ by randomly sampling $m$ 
samples at each node and choosing the best split among the 
$m$ sampled predictors.

\item Make predictions, for the test data based on an 
approximate value $\hat{f}(\vec{z}, D_{n})$, taking a majority 
of votes over the classifiers.

\end{enumerate}
The bootstrapping and ensemble 
scheme adopted by RFCs enables 
them to be 
robust enough to avoid the problem 
of over-fitting
and hence there is no need to 
prune the trees.
For these experiments, the maximum 
depth of each tree was set 
to be 5 and comprised of a forest 
of 20 trees.

\section{Experiments}
\label{experiments}

\subsection{Experiments for the GIM and NIM }

Our influence models were 
evaluated using
three 
synthetic 
networks, SYNTH1, SYNTH2 and 
SYNTH3. SYNTH1 was randomly 
drawn by hand and SYNTH2 
and SYHTH3 were generated from a 
pseudo random number 
generator as in \cite{Matsumoto}. 
All methods were written from 
scratch and 
implemented using Python version 
2.7 (64 bit) on a server with 
8GB of RAM and i3 Processor and 
an average of ten runs 
were taken for each experiments. 
The goal of these 
experiments was to analyze the 
impact of the NIM and 
GIM parameters on the optimal 
solution obtained from 
implementing SDP.

\subsubsection{Dataset 
Description}

We executed our experiments on 
SYNTH1, SYNTH2 and SYNTH3. 
SYNTH1 consisted of 10 nodes, 
SYNTH2 consisted of 2,000 nodes 
and SYNTH3 consisted of 4,500 
nodes.
The SDP method was implemented 
on 
SYNTH1 only, due to 
its complexity. For SYNTH2  
and SYNTH3, the LDH was applied 
and 
values of $p_{0}, 
\alpha$ and $\beta$ were varied. $\beta$ 
was assigned values 
between 0 and 1 whilst for $\alpha$ we 
considered values both less than and 
greater than 1. The results are 
displayed in Figures (\ref{fig:2}- \ref{fig:11}) for experiments involving 5 
impressions over 3 stages.

%
\subsubsection{Sensitivity analysis of the GIM and NIM }

For the sensitivity analysis on 
the GIM 
and NIM their parameter values 
were varied. Figures(\ref{fig:2}-
\ref{fig:5})
display the effect that increasing 
$p_{0}$ 
has on the optimal expected 
number of clicks when 
its value is increased from 
$0.1$ to $0.8$ and $\alpha$ kept constant at 0.25 on SYNTH1, SYNTH2 
and SYNTH3.\newline\indent 
The results indicate that
$p_{0}$ has a significant effect 
on the optimal 
expected number 
of clicks and as $p_{0}$ 
increases so does the 
optimal expected number of 
clicks. This result is not 
surprising since for both the 
NIM 
and GIM  the value for the 
parameter 
$p_{0}$ is additive.
We also note that 
although the optimal expected 
number of clicks increases 
steadily in both the SDP method 
and its heuristic, expected 
click values for $p_{0}$ 
greater than 0.6 increases at a 
greater rate for the 
LDH than the SDP method on all three 
datasets. We believe that this 
is due primarily to the 
construction of the LDH 
algorithm and the structure of 
the synthetic networks. We note that when 
$p_{0}=0.8$, the SDP method generates 
almost 5 clicks under the GIM model 
which demonstrates
significant gains that can be achieved 
by selecting ideal users and suitable 
influence models.\newline\indent
Figures (\ref{fig:6}-\ref{fig:8}), indicate the 
optimal 
expected number of 
clicks on datasets SYNTH1 and SYNTH3, as $\alpha$  
increases
from 0 to 0.9. Figure(\ref{fig:9})
displays the 
results  when $\alpha \geq 1$ on SYHTN1. The 
optimal expected number of clicks 
under 
both the GIM and NIM increases as
$\alpha$ increases. This is expected as $\alpha$ is 
the power in which the GIM is 
raised and is also additive under
the NIM. For a problem 
involving 5 impressions in 3 stages, the 
results in Figure (\ref{fig:9}) ensures 
at 
least 2 clicks with $p_{0}=0.25$. 
That is, at least 
75\% more than the optimal 
expected 
number of clicks generated if all the 
impressions had been placed in 
one stage.
As $\alpha$ increases beyond 5, 
under the NIM and beyond 2 with
the GIM, the optimal expected 
clicks remains constant. 
This result 
is primarily due to the support 
for $p_{k}[i]$:
$0 \leq  p_{k}[i] \leq 1 $ in  
both the NIM and GIM.\newline\indent 
Figure (\ref{fig:10}) and Figure 
({\ref{fig:11}) indicate the 
the optimal expected number of 
clicks as $\beta$ increases. The 
optimal expected number of 
clicks decreases as 
values 
of $\beta$ increases. This is 
expected as in the NIM, the term including 
$\beta$ is being subtracted,  
$(p_{0}+ \alpha \frac{y}{f} - \beta \frac{n}{f})$. 
However we note that at 
some point,
the 
value of the optimal expected 
number of clicks 
remains constant even though 
values of $\beta$ 
continues to increase. This 
result 
is consistently true for graphs 
of 
all sizes. (The results 
illustrating the effect of $\beta$ 
on a graph of 2,000 and 4,500 nodes 
are similar and omitted). \newline\indent
Figures (\ref{fig:4}-\ref{fig:9}) indicate that the 
GIM consistently outperforms NIM 
in generating optimal expected 
number of clicks. These results provide insights into the choice of influence models and role that their parameters play in 
maximizing the expected number of 
clicks and generating revenue for 
the  IM-RO 
problem.

\begin{figure*}
\centering
\begin{multicols}{2}
    \includegraphics[width=\linewidth,height=4cm]{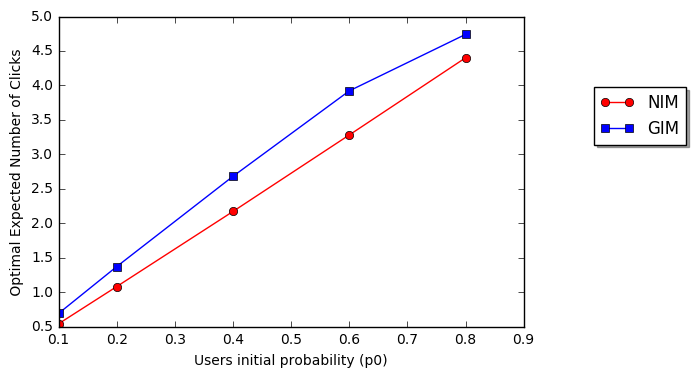}\par 
   \caption{Varying p0 with SDP on SYNTH1}
    \label{fig:2}
   \includegraphics[width=\linewidth,height=4cm]{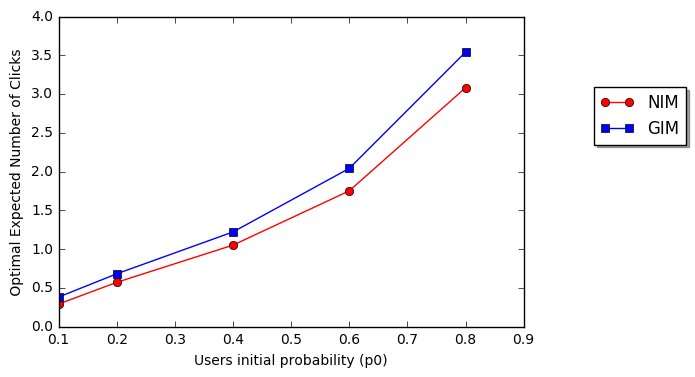}\par 
    \caption{Varying p0 with LDH on SYNTH1} 
    \label{fig:3}
    \end{multicols}
\begin{multicols}{2}
    \includegraphics[width=\linewidth,height=4cm]{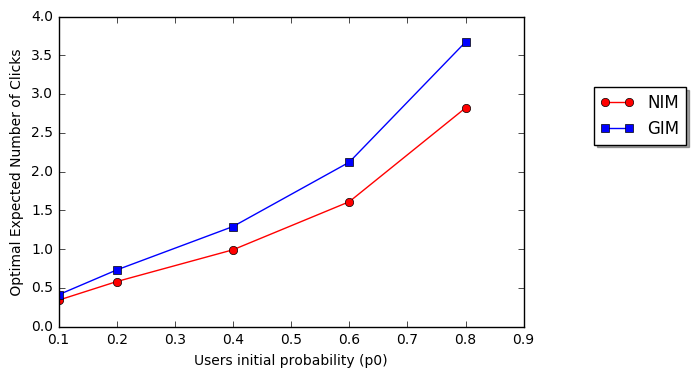}\par
   \caption{Varying p0 with LDH on SYNTH2}
    \label{fig:4}
   \includegraphics[width=\linewidth, height=4cm]{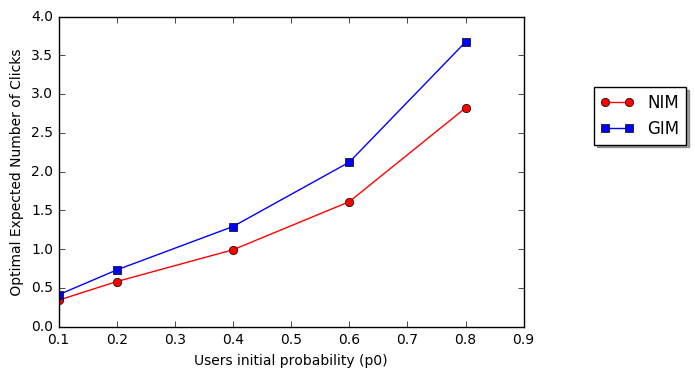}\par 
   \caption{Varying $p_{0}$ with LDH on SYNTH3}
    \label{fig:5}
\end{multicols}

\begin{multicols}{2}
    \includegraphics[width=\linewidth, height=4cm]{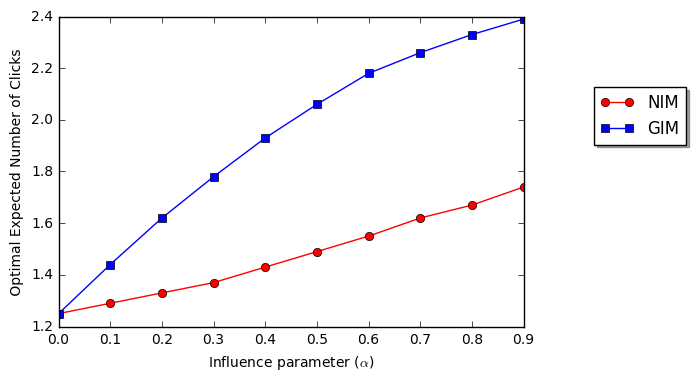}\par 
   \caption{Varying $\alpha$ with SDP on SYNTH1} 
    \label{fig:6}
   \includegraphics[width=\linewidth, height=4cm]{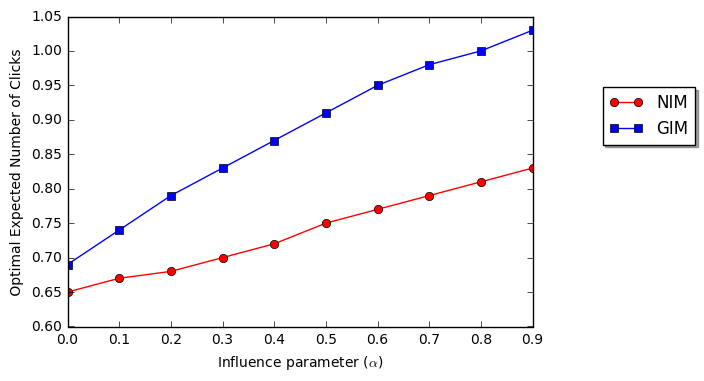}\par 
    \caption{Varying $\alpha$ with LDH on SYNTH1} 
     \label{fig:7}
    \end{multicols}
\begin{multicols}{2}
    \includegraphics[width=\linewidth,height=4cm]{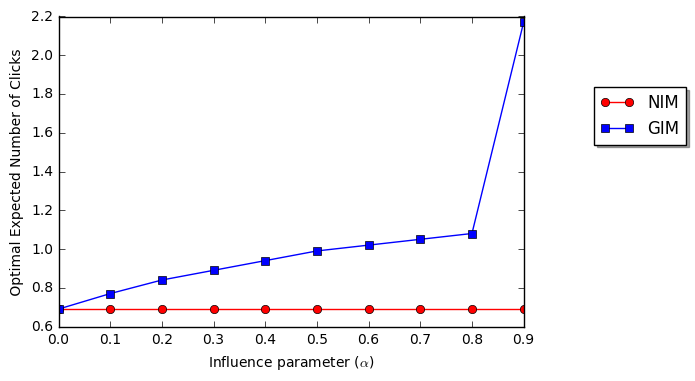}\par
   \caption{Varying $\alpha$ with LDH on SYNTH2}
    \label{fig:8}
   \includegraphics[width=\linewidth, height=4cm]{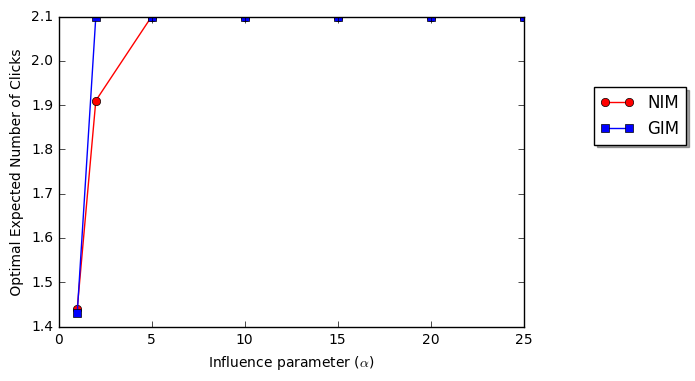}\par
   \caption{Varying $\alpha$ $\geq 1$ with SDP on SYNTH1} 
    \label{fig:9}
\end{multicols}

\begin{multicols}{2}
    \includegraphics[width=\linewidth, height=4cm]{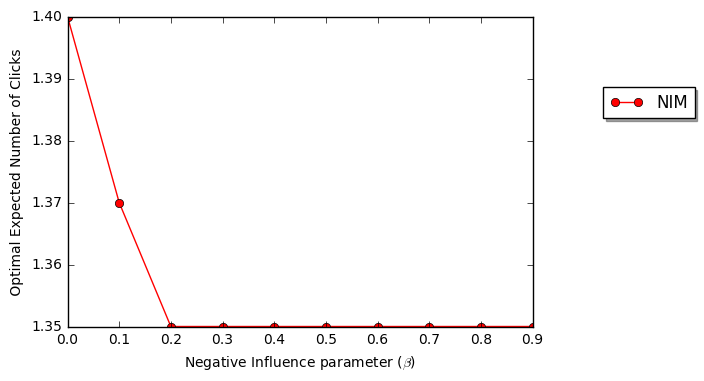}\par
   \caption{Varying $\beta$ with SDP on SYNTH1}
    \label{fig:10}
   \includegraphics[width=\linewidth,height=4cm]{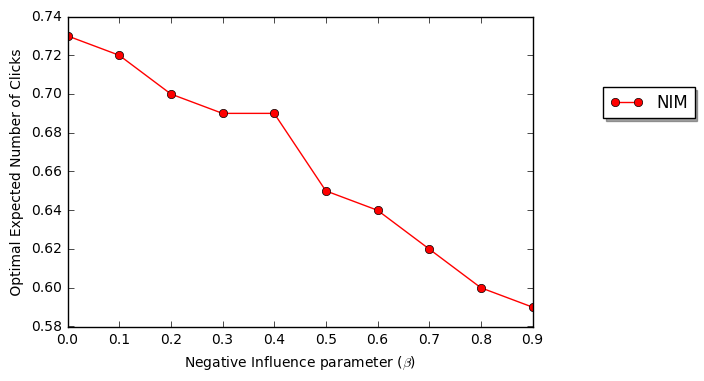}\par
   \caption{Varying $\beta$ $\geq 1$ with LDH on SYNTH1} 
    \label{fig:11}
\end{multicols}
\end{figure*}

\subsection{Estimation of 
$\alpha$}

The Bayesian Hierarchical model was fitted using jags, 
an R interface to JAGS (Just 
Another Gibbs Sampler )\cite{Plummer}, 
which uses Gibbs Sampling to 
estimate the marginal posterior 
distribution for the parameter of 
interest, $\alpha$ in the GIM. The 
MCMC sampling process was allowed 
to simulate for 10,000 iterations 
with a burn-in of 1000 and 
100,000 iterations for a burn-in 
of 10,000 iterations. The process 
involved three chains with the $5-th$ 
iteration in each chain stored 
(thinning). One limitation of the MCMC 
method is that it does not give a clear 
indication of whether it has converged 
\cite{Plummer2}, however, convergence 
was assessed from the trace plots and 
autocorrelation plots. The effective 
accuracy of the 
chain was measured by the Monte Carlo 
standard error
(MCSE) \cite{Kruschke}. To ensure the 
accuracy of the summary statistics we 
provided results in which the MCSE was 
5\% or less, than the posterior 
standard 
deviation \cite{Lunn2}. The results are 
displayed in Tables (\ref{tab:1}-
\ref{tab:5}).

\subsubsection{Microblog Dataset 
for Bayesian Analysis}

With five real-world datasets; 
MICRO0, MICRO1, MICRO2, MICRO3 and
MICRO4 consisting of continuous variables and 
extracted from \cite{Lichman} a 
microblog website, we executed 
our simulations. MICRO0 consisted of 
30,078 
POSTIDs and an average number of 241 
friends, 
MICRO1 consisted 
20,090 POSTIDs and an average 84 
friends, 
MICRO2 consisted of 10,099 POSTIDs and 
an 
average of 21 friends, MICRO3, 6,183 and 
33 friends and MICRO4 consisted of 5,513 
POSTIDs and  an average 
number of 37 friends. We assumed that the average number 
of REPOST was a good indicator of the 
average number of friends in each 
dataset. The overall goal in analyzing these 
datasets was 
to determine an estimate for the 
parameter $\alpha$ in the GIM by utilizing the Bayesian hierarchical 
model.

\subsubsection{Results for Bayesian Analysis}

The results of the experiments  
are summarized in Tables(\ref{tab:1}-
\ref{tab:5}) and 
Figure (\ref{fig:auto}). As seen 
in the Tables, $\alpha$ was 
consistently found to be between 
3.16 and 3.20 with a  prior $\sim 
uniform (0,5)$  
and between 8.15 and 8.22 with a prior 
$\sim uniform (0,10)$  for a burn-in of 10,000 iterations. From 
Table(\ref{tab:1}) and 
Table (\ref{tab:3}), a 
point estimate for $\alpha$ was 
found to be 3.19  with 95 \% CI 
(1.47, 4.91), 3.19 with 95\% 
CI(1.46, 4.92), 3.16 with 95 
\% CI(1.42, 4.91), 3.18 with 
95 \% CI(1.43, 4.91), 3.18 
with 95\% CI(1.44, 4.91) for 
MICRO0, MICRO1, MICRO2, MICRO3 
and MICRO4 respectively for a burn-in of 
10,000 iterations. The 
top part of 
Figure(\ref{fig:auto}) shows  
examples of autocorrelation plots at lag $k$, 
$ACF(k)$, for a burn-in of 1,000 iterations on 
MICRO1 and MICRO2 respectively 
and the bottom two 
parts show examples
of the corresponding 
autocorrelation plots for a burn-
in of 10,000 iterations. One can
see in Figure (\ref{fig:auto})  
from observing the autocorrelation function, that the chains are 
non-autocorrelated, since the 
autocorrelations remain 
particularly close to zero for 
large lags. Not surprisingly, this result is 
further emphasized in the bottom 
two plots of Figure 
(\ref{fig:auto}) when the burn-in 
is 10,000 iterations. Because the 
autocorrelation is an indicator 
of the amount of information 
contained in a given number of 
draws from the posterior, lower 
autocorrelation values are ideal. 
This is also an indication of a 
high level of efficiency or 
mixing of the chains.  
The remaining autocorrelation 
plots were similar to those in Figure 
(\ref{fig:auto}) and therefore 
not included.\newline\indent The MCSE 
is 
similar to the standard error of a 
sample mean and thus, as the sample 
size increases, the standard error 
should also decrease.
Tables (\ref{tab:1}), (\ref{tab:2}) and 
(\ref{tab:4}) display the MCSE for our 
experiments. As seen in Table
(\ref{tab:2}) and Table (\ref{tab:4}),
the Time
Series standard error is the 
smallest on 
MICRO0, the largest dataset, but 
for Table 
(\ref{tab:1}), the standard 
error is the 
smallest on MICRO2. We believe 
that this is due to the insufficient number 
of burn-in 
iterations and its effect on the 
autocorrelation. We note the 
higher autocorrelated values in 
the top two plots of Figure 
(\ref{fig:auto}) hence causing 
an increase in the standard 
error. \newline\indent
In general, we find that the Bayesian 
method is efficient for predicting 
point estimates for $\alpha$, however 
its value
significantly affected by 
the choice of priors. As seen in these 
experiments the point estimate for 
alpha varies greatly when the the 
distribution  of the prior changes. In 
order for us to determine how accurate our 
point estimates are from the true value 
of $\alpha$, a 
dataset comprising of probabilities of
reposting a POST is required.

\begin{table*}
\begin{multicols}{1}
\caption{Shows the empirical mean and standard deviation for $\alpha$ with a  1,000 burn-in, 10,000 iterations and uniform prior (0,5)}
\label{tab:1}       
\begin{tabular}{llllll}
\hline\noalign{\smallskip}
Dataset & Mean & SD &Naive SE& Time Series SE  \\
\noalign{\smallskip}\hline\noalign{\smallskip}
MICRO0 & 3.20348 & 1.05082& 0.01357& 0.01357 \\
MICRO1 & 3.1820 & 1.05897& 0.01369& 0.01420 \\
MICRO2 & 3.19057 & 1.05277& 0.01359& 0.01281 \\
MICRO3 & 3.16701 & 1.04781   & 0.01353 &  0.01353 \\
MICRO4 & 3.13730 & 1.05210& 0.01358& 0.01358 \\
\noalign{\smallskip}\hline
\end{tabular}
\end{multicols}
\begin{multicols}{1}
\caption{Shows the empirical 
mean and standard deviation for 
$\alpha$ with 10,000 burn-in,  100,000 iterations with uniform prior (0,5)}
\label{tab:2}       
\begin{tabular}{llllll}
\hline\noalign{\smallskip}
Dataset & Mean & SD&Naive SE& Time Series SE  \\
\noalign{\smallskip}\hline\noalign{\smallskip}
MICRO0 & 3.193742  & 1.044780& 0.004265  & 0.004265  \\
MICRO1 &3.18722  &1.0461 & 0.004271  &0.004271  \\
MICRO2 & 3.162 & 1.0611 &0.0043  & 0.0043 \\
MICRO3 &3.18184& 1.0518  &0.00429 & 0.00429 \\
MICRO4 & 3.1758  &1.0517  & 0.00429  & 0.004294 \\
\noalign{\smallskip}\hline
\end{tabular}
\end{multicols}
\begin{multicols}{1}
\centering
\caption{Shows the quartiles for $\alpha$ with uniform prior (0,5)}
\label{tab:3}       

\begin{tabular*}{\linewidth}{@{\extracolsep{\fill}}p{0.10\linewidth}p{0.05\linewidth}p{0.06\linewidth}p{0.06\linewidth}p{0.06\linewidth}p{0.06\linewidth}p{0.12\linewidth}@{}}

\toprule
Dataset &Update & 2.5\% & 25\%  & 
50\% & 75\% & 97.5\%  \\
\midrule
MICRO0 & 1000& 1.470  & 2.283 &3.210  & 4.121& 4.903 \\
MICRO0 & 10000&1.470 &2.288 & 3.193 &4.102 & 4.908  \\
MICRO1 & 1000&1.437  & 2.255  & 3.194 & 4.075 & 4.919 \\
MICRO1 & 10000& 1.457 & 2.283  &3.184  & 4.089 & 4.908  \\
MICRO2 &1000 &1.437 & 2.271  & 3.243 & 4.092& 4.900 \\
MICRO2 & 10000& 1.421 & 2.235 & 3.155& 4.083 & 4.912  \\
MICRO3 & 1000&1.443  & 2.244  & 3.193 & 4.079 & 4.884 \\
MICRO3 & 10000& 1.431 &2.276 & 3.182  &4.095 & 4.910  \\
MICRO4 & 1000& 1.430 & 2.225 & 3.140& 4.041 & 4.882 \\
MICRO4 & 10000&1.438 & 2.270 &3.176 &4.085 & 4.912  \\
\bottomrule
\end{tabular*}
\end{multicols}
\begin{multicols}{1}
\caption{Shows the empirical 
mean and standard deviation for 
$\alpha$ with 10,000 burn-in,  100,000 iterations with uniform prior (0,10), }
\label{tab:4}       
\begin{tabular}{llllll}
\hline\noalign{\smallskip}
Dataset & Mean & SD&Naive SE& Time Series SE  \\
\noalign{\smallskip}\hline\noalign{\smallskip}
MICRO0 & 8.21104 & 3.91881 & 0.016 & 0.01549 \\
MICRO1 & 8.17812 &3.91684 & 0.01599 &0.01617 \\
MICRO2 & 8.15377 & 3.93972 &0.01608 & 0.01582  \\
MICRO3 &8.18965 & 3.92468 & 0.01602&  0.01564 \\
MICRO4 & 8.19130 & 3.91978 & 0.01600 & 0.01585 \\
\noalign{\smallskip}\hline
\end{tabular}
\end{multicols}
\begin{multicols}{1}
\centering
\caption{Shows the quartiles for $\alpha$ with uniform prior (0,10)  }
\label{tab:5}       

\begin{tabular*}{\linewidth}{@{\extracolsep{\fill}}p{0.10\linewidth}p{0.05\linewidth}p{0.06\linewidth}p{0.06\linewidth}p{0.06\linewidth}p{0.06\linewidth}p{0.10\linewidth}@{}}

\toprule
Dataset &Update & 2.5\% & 25\%  & 
50\% & 75\% & 97.5\%  \\
\midrule

MICRO0 & 10000&1.734 &4.829 &8.198&11.583 &14.658  \\

MICRO1 & 10000& 1.694 &4.791 & 8.199& 11.528 & 14.684  \\

MICRO2 & 10000& 1.67 & 4.742& 8.154 &11.549 & 14.675  \\

MICRO3 & 10000& 1.681 &4.821 &8.227& 11.552& 14.655  \\

MICRO4 & 10000& 1.682 &4.820 & 8.242 & 11.527 & 14.666   \\
\bottomrule
\end{tabular*}
\end{multicols}
\end{table*}

\begin{figure*}
\begin{multicols}{2}
    \includegraphics[width=\linewidth, height=3.5cm]{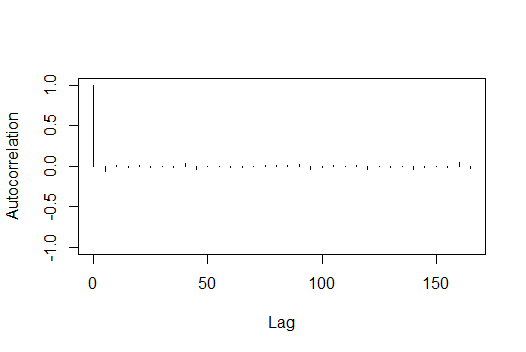}\par 
 \includegraphics[width=\linewidth, height=3.5cm]{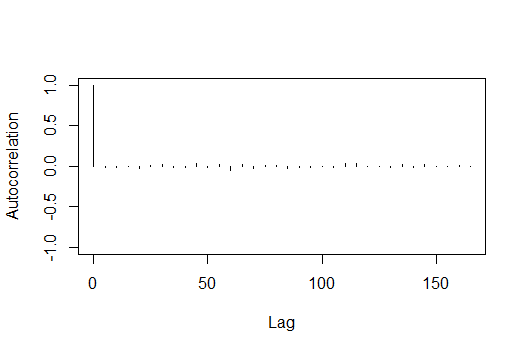}\par 
 
    \end{multicols}
\begin{multicols}{2}
    \includegraphics[width=\linewidth, height=3.5cm]{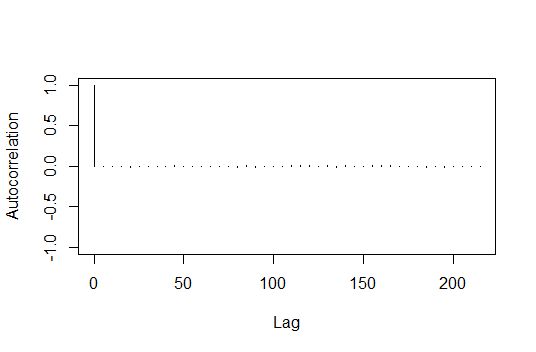}\par
   
  \includegraphics[width=\linewidth, height=3.5cm]{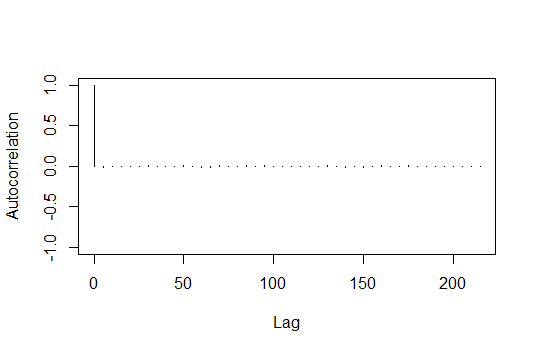}\par

\end{multicols}
\caption{Autocorrelation plots for $\alpha$ with prior $\sim$ (0,5)  }
\label{fig:auto}
\end{figure*}

\subsection{Estimation of 
$p_{0}$}

We conducted experiments on three 
datasets, two of which were extracted 
from the OSN, Twitter, and
obtained in \cite{LeskovecS} 
and the third, a 
microblog dataset obtained 
from \cite{Lichman}. All methods were 
implemented through Apache Spark MLib 
package \cite{Meng} with Scala  
version 2.1.0 on a server with 
8GB of RAM and i3 Processor.  
An average of ten runs 
were taken for each experiments. 
The objective of these 
experiments was to obtain the most
efficient algorithm for classifying 
and predicting the data in order 
to obtain an accurate estimate 
for the parameter $p_{0}$,  a user's initial probability of clicking on an 
impression, in the absence of any influence from 
friends. Our approach is based on modeling  $p_{0}$ as a function,  $p_{0}$ :V $\rightarrow$ 
[0,1] and implementing the DTC, NCB and  RFC
algorithms in order to learn this parameter based on features from three datasets.

\subsubsection{Dataset Description}
The three datasets, TWITT1, TWITT2 
and MICRO5 entailed 
nominal and binary features and a 
class 
label consisting of two outcomes 
or 
classes corresponding to a user 
tweeting 
or not tweeting a phrase. TWITT1 
consisted of 16 features and 447 
Instances, MICRO5 consisted of 3 
features and 142,369 instances while 
TWITT2 consisted on 10 features and 179 
instances. 
TWITT1 and TWITT2 were made up 
of binary features whilst 
MICRO5 consisted of nominal 
features. 
A detailed description of the 
features 
for each dataset can be found 
at \cite{Leskovec}. 
For further analysis we divided each dataset 
into ten disjoint training and test sets as follows:
\vspace*{-\baselineskip}
\begin{itemize}
\item 10\% training and 90\% test data.
\item 20\% training and 80\% test data.
\item 30\% training and 70\% test data.
\item 40\% training and 60\% test data.
\item 50\% training and 50\% test data.
\item 60\% training and 40\% test data.
\item 70\% training and 30\% test data.
\item 80\% training and 20\% test data.
\item 90\% training and 10\% test data.
\end{itemize}

\subsubsection{Performance Measure}

For an analysis on the performance 
of the DTC, NBC and RFC 
algorithms, the receiver operating 
characteristics (ROC) was used. This is a 
plot of $p(\hat{y}=1|y=1)$, known 
as the true positive rate (TPR) 
against 
$p(\hat{y}=1|y=0)$, the false 
positive rate (FPR) in a 
function which is a fixed 
threshold for a parameters $\tau$ 
is used.
\newline\indent
The quality of the ROC curve is 
summarized by a 
single number 
using 
the area under curve, $AUC$. The 
$ROC$ sensitivity ranges 
from 0 to 1 with higher AUC scores being 
preferred. In 
general, a more accurate 
classifier 
has a AUC value closer to 1 and 
very 
low AUC values indicate that the 
classifier is possibly finding a 
relationship 
with the data that is exactly the 
opposite than what is expected.   
 \newline\indent
Another metric used to evaluate 
the 
performance 
of the algorithm in these 
experiments was accuracy, in other 
words 
analyzing whether a prediction was 
correct or not. This metric 
however 
can be 
misleading since 
its prediction are based primarily 
on the 
datasets used. For example a 
predictive model can be evaluated 
as 
being 
the 90\% accurate simply because 
90\% of the data used belonged to 
one class. Figures ((\ref{data1}-
\ref{data3}) was also used to determine the accuracy of each algorithm.

\begin{figure*}
\begin{multicols}{2}
 \includegraphics[width=\linewidth]
   {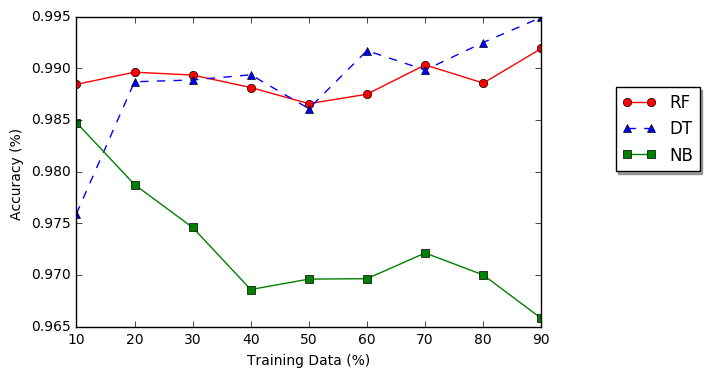}\par
\centering
 \caption{447 Instances, 16 features, 2 classes}
 \label{data1}
 \includegraphics[width=\linewidth]
   {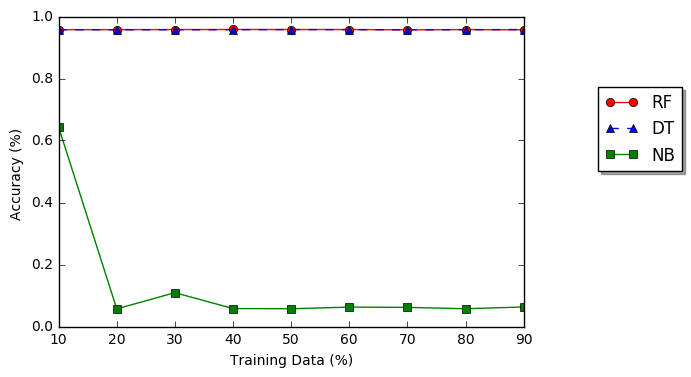}\par
\centering
 \caption{142,369 Instances 3 features , 2 classes.}
 \label{data2}
 \end{multicols}

\begin{multicols}{1}
 \includegraphics[width=\linewidth]
   {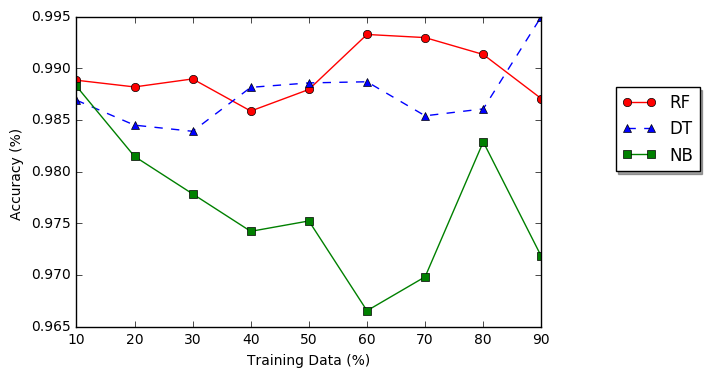}


\end{multicols}
 \caption{179 Instances, 10 features, 2 classes.}
  \label{data3}
\end{figure*}

\hfill \break
\newpage
\begin{table*}[t]
\caption{A comparison of the various algorithms in terms of metrics}
\label{comparison}       
\begin{tabular*}{\linewidth}{@{\extracolsep{\fill}}p{0.15\linewidth}p{0.1\linewidth}p{0.1\linewidth}p{0.1\linewidth}p{0.1\linewidth}p{0.1\linewidth}p{0.1\linewidth}@{}}

\toprule
Algorithm && AUC  && & Accuracy  \\
& TWITT1& MICRO5& TWITT2 & TWITT1 & MICRO5 & TWITT2 \\

\midrule
DTC & 0.865  & 0.755 & 0.977&0.989& 0.958 & 0.989\\
NBC & 0.613 & 0.794 & 0.621 & 0.973& 0.057 & 0.966\\
RFC & 0.907  & 0.856 & 0.977& 0.989 &0.958 &0.989\\

\bottomrule
\end{tabular*}
\end{table*}

\begin{table}

\caption{Average time for running algorithms (msec) over entire datasets}
\label{runningtimes}       
%
%
\begin{tabular*}{\linewidth}{@{\extracolsep{\fill}}p{0.15\linewidth}p{0.1\linewidth}p{0.1\linewidth}p{0.1\linewidth}p{0.1\linewidth}@{}}
\toprule
Dataset & RFC& DTC & NBC \\
\midrule
TWITT1 & 6000  & 7000 & 500
  \\
MICRO5 & 11000 & 8000& 3000 \\
TWITT2 & 8000  & 8000 & 2000 \\

\bottomrule
\end{tabular*}
\end{table}

\begin{table}

\caption{Probability of predicting class 1, based on 100 samples }
\label{probabilityprediction}       
%
%
\begin{tabular*}{\linewidth}{@{\extracolsep{\fill}}p{0.15\linewidth}p{0.1\linewidth}p{0.1\linewidth}p{0.1\linewidth}p{0.1\linewidth}@{}}
\toprule
Dataset & RFC& DTC & NBC \\
\midrule
TWITT1 & 0.01   & 0.004 & 0.02 
  \\
MICRO5 & 0.97 & 0.99 & 0.00 \\
TWITT2 & 0.02  & 0.01 & 0.03 \\

\bottomrule
\end{tabular*}

\end{table}

\subsection{Results for Machine Learning Algorithms}

The results in Figure 
(\ref{data1}), Figure 
(\ref{data2}) and Figure 
(\ref{data3}) confirm the 
original 
hypothesis and 
work done in \cite{Caruana} that 
the RFC 
outperforms the NBC and DTC 
algorithms in terms 
of accuracy. Its accuracy is the 
best in all 
three datasets and it is clear 
that the RFC learns 
faster than both the DTC and NBC as 
the 
prediction accuracy of the RFC is higher than NBC
and 
DTC when a
small percentage of the training data is used, only 10\%.
\newline\indent
Table (\ref{comparison}) shows the 
results 
for the algorithms and their 
respective AUC 
and accuracy. It is worth noting 
that the NBC 
algorithm lags considerably behind 
the RFC and 
DTC based on evaluations on both 
metrics. We 
also note that for MICRO5 when 
evaluated by 
the accuracy metric, has smaller 
values. We believe 
that this result is due to the 
limited number of 
features used in proportion to the 
size of the dataset. Table 
(\ref{comparison}) 
also shows that the RFC algorithm 
outperforms the NBC and DTC 
algorithm in terms of accuracy and 
AUC.
\newline\indent
Table (\ref{runningtimes}) 
displays the 
running times for the NBC, DTC and 
RFC algorithms 
on all three datasets.
Despite having the worst 
performance in terms accuracy and 
AUC when compared to the DTC and 
RFC algorithms, the running times 
of the NBC algorithm is 
considerably less 
than the 
runningtimes for the DTC and RFC 
as the NBC converges towards 
asymptotic accuracy at a faster 
rate.
\newline\indent
The experimental results 
displayed in Table 
(\ref{probabilityprediction}), 
indicate the average 
probability of predicting class 1 (the probability of tweeting) 
for each classifier on 
each dataset. We conclude that the 
most accurate probability for 
TWITT1 is 0.01, predicted by 
the RFC, the DTC predicts a 
probability of 0.004 and we 
believe that  this value is due 
primarily due to over-fitting of 
the DTC. For MICRO5, users had a 
much higher probability of 
predicting the phrase, as the best  
probability was selected as 0.99 determined by 
the RFC algorithm due to its AUC and accuracy value. The NBC 
algorithm performs considerable 
poorly for this case predicting a 
value of 0. Again, we attribute 
this result to the limited number 
of features used in the MICRO5 dataset and 
the performance of 
NBC algorithm. In general, we find that the probability of retweeting a character or phrase depended significantly on the dataset. \newline\indent 
The results demonstrate that an estimate for $p_{0}$ can be easily obtained by using supervised learning algorithms through Apache Spark.
They also implicitly provide additional insights for advertisers to 
achieve considerable gains by 
spreading 
information or advertising products.

\subsection{Cross Validation}

N fold Cross-Validation was 
implemented as a technique 
which determined the 
best model for each dataset by 
training and testing 
the model on different portions 
of the datasets. 
The idea behind the technique 
involves splitting the 
dataset into N folds then for 
each fold  $\thinspace$
$n $
$\in {1,2,3,...,N}$, 
the model is trained on 
all but the $nth$ fold and tested on the $nth$ fold in 
a robin-robin fashion. Cross validation has proven 
to be an effective procedure for removing the bias 
out of the apparent error rate and has been 
implemented 
in numerous papers 
\cite{Lachenbruch,Krstajic,Simon,Stone,Kohavi,Dudoit,Sylvain}
\newline\indent
Table (\ref{crossvalidation}) 
displays the results 
for the best model determined by 5 
fold cross 
validation. The technique computes 
the average error 
over all 5 folds and uses it as a 
representative for 
the error in the test data. The best 
model achieved 
through the cross validation process 
was then evaluated using the AUC 
metric.

\begin{table}[H]

\centering
\caption{Evaluation of 5 fold Cross Validation using AUC metric}
\label{crossvalidation}       
%
%
\begin{tabular}{p{2cm}p{2cm} p{2cm}}
\hline\noalign{\smallskip}
Dataset & DTC &  RFC \\
\noalign{\smallskip}\hline\noalign{\smallskip}
TWITT1 & 0.865 &     0.952
   \\
 MICRO5 & 0.755 &  0.874  \\
TWITT2 & 0.977 &  0.977\\
 \noalign{\smallskip}\hline\noalign{\smallskip}
\end{tabular}

\end{table}

The results in Table 
(\ref{crossvalidation}) indicate 
an evaluation of the best model using 
the RFC and DTC and learning 
algorithms achieved through 5 fold 
cross validation 
and evaluated by the AUC metric. We can conclude that the 
RFC algorithm has the smallest 
error rate when evaluated using 
the AUC metric. For TWITT2, 
the DTC and RFC were both proven 
to be ideal for predicting 
$p_{0}$  however the RFC 
performed consistently well in 
all three datasets and can be 
considered as an effective 
method for obtaining $p_{0}$.



\section{Conclusion}
\label{conclusion}

In this paper, we have presented a 
novel analysis on the influence models 
for the IM-RO problem. The IM-RO 
problem was first formally defined in 
\cite{Lawrence2} as a novel approach 
to the well-known IM problem which 
diverted from the theory of submodular 
functions and focused on maximizing
expected gains for the 
advertiser. This approach is achieved 
through implementing SDP and is 
demonstrated to have lucrative gains 
when evaluated with the GIM and NIM  on 
various real and synthetic networks. We 
have shown how the composition of the  
GIM and NIM and varying their 
parameters  affect the optimal expected 
number of clicks generated.
Our results show that the influence 
models as well as the structure of the 
OSN play an integral role in 
optimizing 
clicks and ultimately generating 
revenue to the advertiser. We 
have also 
introduced a Bayesian and 
Machine 
Learning approach for estimating the parameters $\alpha$ and 
$p_{0}$ of the GIM which 
is easily implementable in the 
standard 
BUGS and Apache Spark softwares, 
respectively. Results indicate that the value for $p_{0}$ relies heavily on the particular character or phrase being retweeted and that the RFC is the most efficient algorithm for computing $p_{0}$. 
\newline\indent 
There are several directions for future 
work. First, we
would like to apply the methods
to real
datasets for which
knowledge of a user's  probability of making a purchase 
at specific intervals in 
time, is 
available. That is, we would like to apply 
our machine learning algorithms 
to determine 
$p_{0k}$, and a user's 
probability 
of 
clicking on an impression
with the knowledge of whether 
or 
not their friends have clicked 
on the 
impression at all stages, 
$k$. This will enable us to 
further 
explore our Bayesian analysis. 
Our 
results indicate that the
point estimates of $\alpha$ are
significantly affected by the 
choice of 
priors. Hence we will be able 
to determine how accurate our 
estimates of $\alpha$ are from its 
true value and make more 
informed 
decisions about the choice of 
priors. Second, we would like to
further explore influence models 
for the IM-RO problem in order 
to improve on the optimal 
expected number of clicks 
generated. Third, we would like 
to investigate alternative data 
science techniques for obtaining 
the parameters of these 
influence models. By defining a 
likelihood function on the 
parameters of an influence 
model, techniques such as the EM 
algorithm \cite{McLachlan} can be implemented to 
obtain the optimal set of 
parameter 
values.

\end{document}